\pdfoutput=1
\documentclass[a4paper,twoside]{article}

\usepackage{float}
\usepackage{subcaption}
\usepackage{amsmath}
\usepackage{amsfonts}
\usepackage{tabularx}
\usepackage[hidelinks]{hyperref}
\usepackage{booktabs}
\usepackage{setspace}
\usepackage{graphicx,array,xurl,placeins}

\usepackage{epsfig}
\usepackage{calc}
\usepackage{amssymb}
\usepackage{amstext}
\usepackage{amsthm}
\usepackage{multicol}
\usepackage{pslatex}
\usepackage{apalike}
\usepackage{algorithm2e}
\usepackage[bottom]{footmisc}
\usepackage{SCITEPRESS} 

\renewcommand\orcidAuthor[1]{
\hspace*{-1mm}\includegraphics[width=0.3cm]{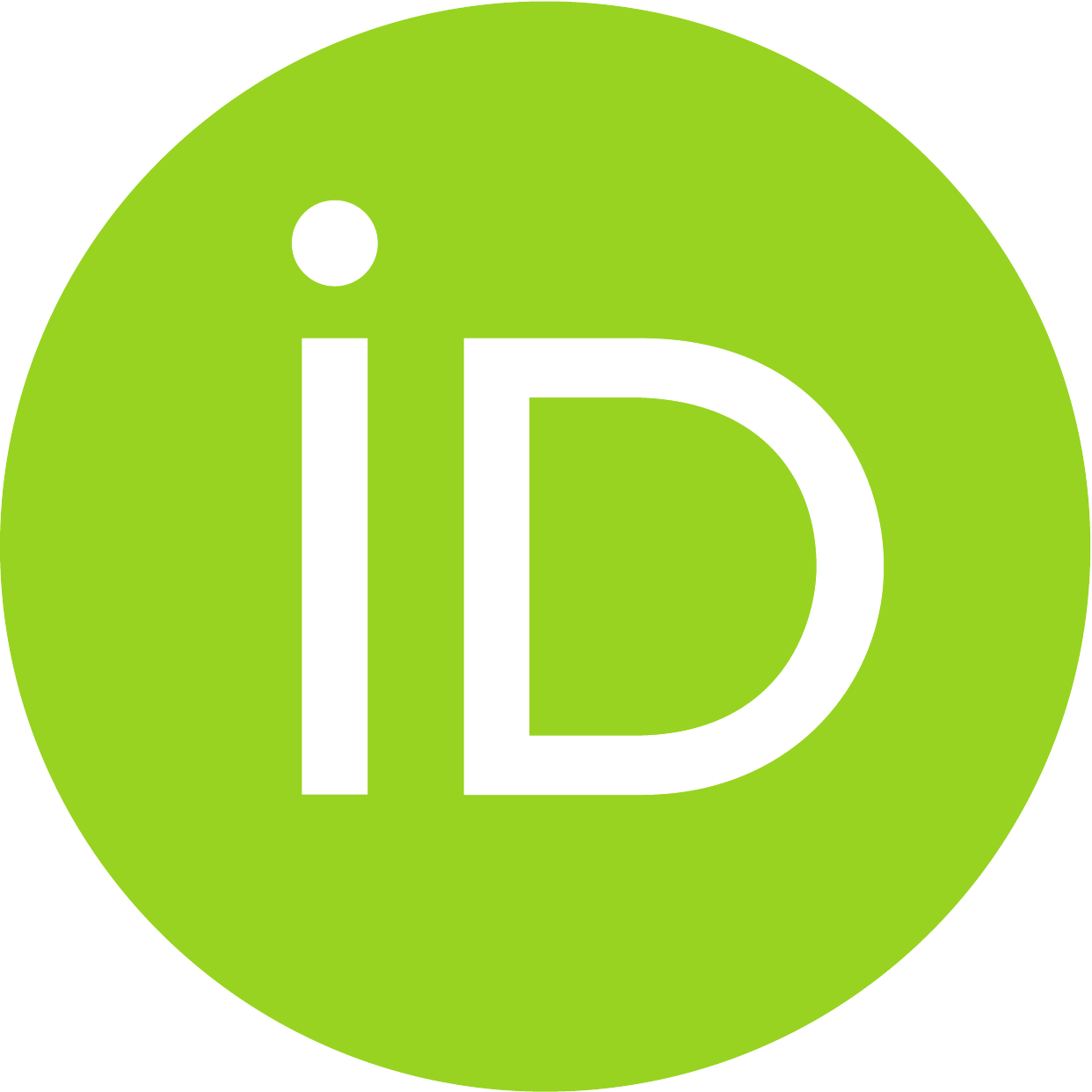}\thanks{\protect\includegraphics[width=0.3cm]{orcid.pdf}~https://orcid.org/#1}\hspace*{-1mm}
}

\makeatletter
\newcommand{\citet}[1]{%
  \begingroup
    \def\@cite##1##2{##1}%
    \cite{#1},%
  \endgroup
}
\makeatother

\begin{document}

\title{BladeR: Scalable, Shape-Guided 3D Reconstruction of Wind Turbine Blades from UAV Imagery}

\author{
\authorname{Jonathan Sterckx\sup{1}\orcidAuthor{0009-0001-1895-9549}, Michiel Vlaminck\sup{1}\orcidAuthor{0000-0003-3986-823X} and Hiep Luong\sup{1}\orcidAuthor{0000-0002-6246-5538}}
\affiliation{\sup{1}IPI-URC-imec, Ghent University, Sint-Pietersnieuwstraat 41, Ghent, Belgium}
\email{\{jonathan.sterckx, michiel.vlaminck, hiep.luong\}@ugent.be}
}


\abstract{The structural integrity of wind turbine blades is critical for safe, efficient energy production. Accurate 3D reconstruction allows precise localization and quantification of wear and early damage detection. While drone-based inspections are standard, their high-resolution imagery poses challenges such as weak texture, lighting variability, and high computational cost. We present BladeR, a scalable pipeline for high-quality 3D reconstruction from inspection imagery. A hierarchical sparse reconstruction strategy leverages GPS/IMU data and blade shape priors to guide advanced keypoint matchers on full-resolution images. For dense reconstruction, a tiled Gaussian Splatting framework processes 45 MP images at native resolution, isolates the blade from background, and adaptively enhances detail recovery along the leading edge. Validation against commercial and open-source baselines shows significant improvements in model completeness and detail.}

\onecolumn
\maketitle
\normalsize
\setcounter{footnote}{0}
\vfill

\begingroup
\renewcommand\thefootnote{}
\footnotetext{\footnotesize This is the authors' version of the paper published in \emph{Proceedings of the 21st International Conference on Computer Vision Theory and Applications (VISAPP 2026) -- Volume~3}, pages 119--130. DOI: \href{https://doi.org/10.5220/0014309700004084}{10.5220/0014309700004084}. Paper published under CC license (CC BY-NC-ND 4.0). Proceedings Copyright \copyright~2026 by SCITEPRESS -- Science and Technology Publications, Lda.}
\addtocounter{footnote}{-1}
\endgroup


\section{INTRODUCTION}

Wind energy is a critical component of the global transition to renewable energy \cite{gwec2024globalwind}. The operational efficiency and lifespan of wind turbines depend heavily on blade integrity. Leading edge erosion, for example, degrades aerodynamic performance, reduces energy output and shortens blade lifespan \cite{herring2019increasing}, \cite{ozccakmak2024determination}, making regular inspections essential \cite{shafiee2019maintenance}.

Unmanned Aerial Vehicles (UAVs) have revolutionized turbine inspections by providing safer, faster, and more cost-effective alternatives to manual methods \cite{poleo2021estimating}. While 2D drone images are useful for detecting surface defects \cite{xu2019wind}, \cite{memari2024review}, full 3D reconstructions offer richer insights by enabling precise volumetric quantification, defect localization, and differentiation between true damage and surface artifacts \cite{sterckx2025segmentation}. They also support defect tracking and blade deformation analysis, enabling more comprehensive digital twin models, with room to advance beyond current 2D techniques \cite{hu2025digital}.

Prior studies demonstrate the feasibility of photogrammetry for blade reconstruction \cite{nielsen2020high}, \cite{zhang2020quantifying}, but often rely on decommissioned blades under controlled conditions, without a scalable pipeline. \citet{sterckx2025accurate} proposed a custom reconstruction pipeline with improved accuracy, yet it does not leverage domain-specific knowledge and its applicability remains limited to small datasets.

To address these limitations, we present a pipeline tailored to real-world inspections and validated on a new UAV dataset. Our contributions are:
\begin{itemize}
	\item \textbf{A high-overlap UAV dataset} of two operational blades captured with a 45 MP camera under varied conditions.
	\item \textbf{A scalable preprocessing pipeline} that uses automated segmentation and blade-centered tiling to enable full-resolution image processing without exceeding memory limits.
	\item \textbf{A robust two-stage sparse reconstruction} that leverages GPS data and a blade-geometry prior to concentrate keypoint matching along the leading edge, reducing the number of required image-matching operations.
	\item \textbf{A custom Gaussian Splatting framework} with soft background masking, blade-specific adaptive densification, and tiled full-resolution processing, allowing reconstruction at resolutions significantly higher than those supported by standard dense reconstruction methods.
\end{itemize}

Our method produces more complete and detailed 3D models than leading commercial and open-source software, while scaling effectively to high-resolution imagery and large inspection datasets—providing a solid foundation for improved damage assessment and maintenance planning.


\section{RELATED WORK}

Only a few works have addressed image-based 3D reconstruction of wind turbine blades.  

\citet{zhang2020quantifying} studied how lighting and image blur affect UAV-based photogrammetric reconstruction. Their experiments used a decommissioned blade section indoors, combining natural and artificial textures. 3D models were generated with Agisoft Metashape \cite{metashape} and compared against a high-precision 3D-scanned reference.  

\citet{nielsen2020high} performed a feasibility study using high-resolution SfM and MVS with Agisoft Metashape to evaluate leading-edge roughness on a decommissioned blade tip with simulated erosion. Their data was collected outdoors with a handheld camera.  

\citet{hung2025three} explored a multimodal approach, capturing visible and thermal images of a small rooftop turbine using UAVs. They applied NeRF \cite{mildenhall2021nerf} to visible imagery and projected thermal views onto the geometry to localize defects. While promising for multimodal inspection, the resulting 3D models lacked the accuracy needed for precise defect measurement.  

\citet{sterckx2025accurate} proposed a custom reconstruction pipeline outperforming off-the-shelf tools in accuracy and robustness. They introduced hierarchical keypoint tracking to handle low-texture regions and variable illumination, and a Gaussian Splatting variant with foreground isolation for dense reconstruction. However, their approach was limited to small datasets, relied on computationally intensive matching, required downscaling (losing critical detail) and did not incorporate blade-specific enhancements.  

To overcome these limitations, this paper introduces a high-overlap UAV dataset of an operational blade under realistic conditions and an enhanced pipeline with foreground-aware tiling, blade geometry-informed matching, and blade-specific dense reconstruction improvements, achieving greater scalability and fidelity.


\section{PRELIMINARIES}

3D reconstruction from multi-view images typically involves two stages: sparse reconstruction using Structure from Motion (SfM), and dense reconstruction via Multi-View Stereo (MVS) or newer alternatives like Gaussian Splatting (GS).

\begin{figure*}[ht]
	\centering
	\includegraphics[width=0.99\linewidth]{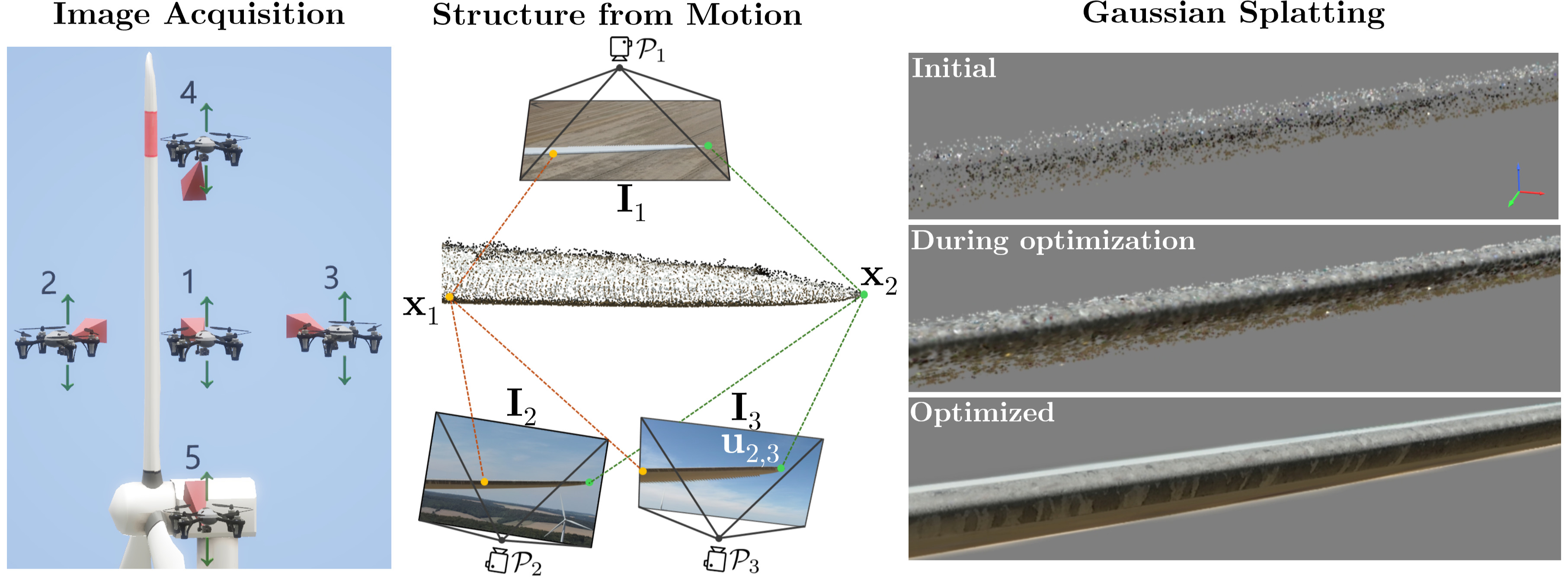}
	\caption{Multi-view 3D reconstruction. Images are acquired from multiple viewpoints. SfM reconstructs sparse 3D points $\mathbf{x}_i$ and camera poses $\mathcal{P}_j$ from keypoint matches $\mathbf{u}_{i,j}$ across images $\mathbf{I}_j$. GS initializes Gaussians on the recovered points and iteratively refines their parameters to align with the input images.}
	\label{fig:high_level_pipeline}
\end{figure*}

\subsection{Sparse 3D Reconstruction}

SfM reconstructs a sparse 3D point cloud and camera poses from overlapping images \cite{SfM_survey}, as illustrated in Fig. \ref{fig:high_level_pipeline} (middle). An incremental SfM pipeline \cite{colmap} typically involves:
\begin{enumerate}
	\item \textbf{Keypoint detection and description:} Image features are extracted using algorithms such as SIFT \cite{sift} or ORB \cite{orb}, which encode the appearance of local image regions into high-dimensional descriptors.
	
	\item \textbf{Keypoint matching:} Keypoint descriptors are matched across image pairs, using nearest-neighbor search or neural networks \cite{sarlin2020superglue}, \cite{lindenberger2023lightglue}. Geometric verification is applied to discard outlier matches.
	
	\item \textbf{Incremental reconstruction:} Starting from an image pair, their relative pose is estimated and corresponding points are triangulated. Additional images are registered via PnP \cite{pnp}, extending the cloud. Outliers are rejected via RANSAC \cite{ransac}.
	
	\item \textbf{Bundle adjustment:} Global optimization refines 3D points and camera poses \& intrinsics to minimize reprojection error \cite{bundle_adjustment}.
\end{enumerate}

Popular open-source SfM systems include COLMAP \cite{colmap}, while Agisoft Metashape \cite{metashape} is a widely used commercial alternative. Recent methods such as Pixel-Perfect SfM \cite{pixel_perfect_sfm}, Detector-Free SfM \cite{detectorfreesfm}, MASt3R \cite{leroy2024grounding}, and VGGT \cite{wang2025vggt} leverage deep learning for improved accuracy, robustness, and speed, but their reliance on neural networks limits scalability to high-resolution imagery. This work introduces a hybrid approach to overcome these scalability challenges, integrating traditional SfM with deep learning-based matchers using a tile-based strategy.

\subsection{Dense 3D Reconstruction}

Dense reconstruction generates detailed surfaces from the input images and sparse reconstruction outputs. Traditionally, Multi-View Stereo (MVS) computes dense depth maps by matching pixels across multiple views and then fuses the depth maps into a detailed 3D representation \cite{MVS_survey}.

More recently, Gaussian Splatting (GS) \cite{kerbl20233d} models the scene with 3D Gaussians initialized on SfM points. These Gaussians are iteratively optimized in location, shape, and opacity so that renderings match input images (illustrated in Fig. \ref{fig:high_level_pipeline}, right), using a combination of $\mathcal{L}_1$ and $\mathcal{L}_\text{SSIM}$ losses with differentiable rendering. Adaptive Gaussian splitting, cloning, and pruning balances detail and memory usage.

By default, GS minimizes photometric error without enforcing geometric constraints, which can cause artifacts in low-coverage regions (“floating” or “needle” Gaussians) \cite{mip_splatting}. Extensions address these issues: Mip-Splatting \cite{mip_splatting} applies 3D low-pass filtering; Deblur-GS \cite{deblur_gs} and BAD-Gaussians \cite{bad_gaussians} improve robustness to defocus and motion blur; 2DGS \cite{2dgs} uses surface-oriented Gaussians with novel regularization.

Compared to MVS, GS is faster and provides a flexible framework to handle challenging imaging conditions such as blur and lighting variation.
Unlike NeRF \cite{nerf}, which encodes a 3D scene implicitly in the weights of a neural network and generates novel views by querying the network with a 5D vector (position and viewing direction), GS offers an explicit representation, making the underlying structure directly accessible. This facilitates tasks such as defect localization and surface analysis, making it particularly well-suited for wind turbine blade inspection.


\section{METHODS}

\begin{figure*}[ht]
	\centering
	\includegraphics[width=0.9\linewidth]{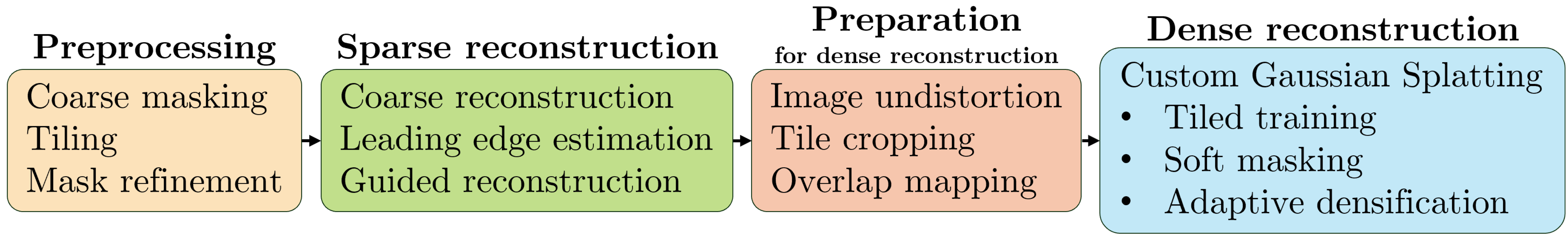}
	\caption{High-level overview of \textit{BladeR}, our proposed 3D reconstruction pipeline.}
	\label{fig:high_lvl_diagram}
\end{figure*}

This section describes the data collection process and our 3D reconstruction pipeline, \textit{BladeR}, covering data preprocessing as well as sparse and dense reconstruction stages. Figure~\ref{fig:high_lvl_diagram} provides a high-level overview of \textit{BladeR}, illustrating the workflow from preprocessing through to dense 3D reconstruction.

\subsection{Data Collection}

To develop and evaluate a 3D reconstruction pipeline for real-world inspections, we acquired high-overlap UAV images of two operational wind turbine blades using a DJI Zenmuse P1 camera (45 MP, 35 mm focal length) mounted on a drone with integrated GPS and IMU sensors.

Accurate 3D reconstruction requires multiple viewpoints, but dense coverage is impractical due to drone battery limits and inspection time. Since wear mainly occurs along the leading edge, we fly multiple lines capturing it from a frontal angle (angle 1 in Fig.~\ref{fig:high_level_pipeline}, left), azimuthal angles (2 and 3), and polar oblique angles (4 and 5).

\begin{figure*}[ht]
	\centering
	\includegraphics[width=0.99\linewidth]{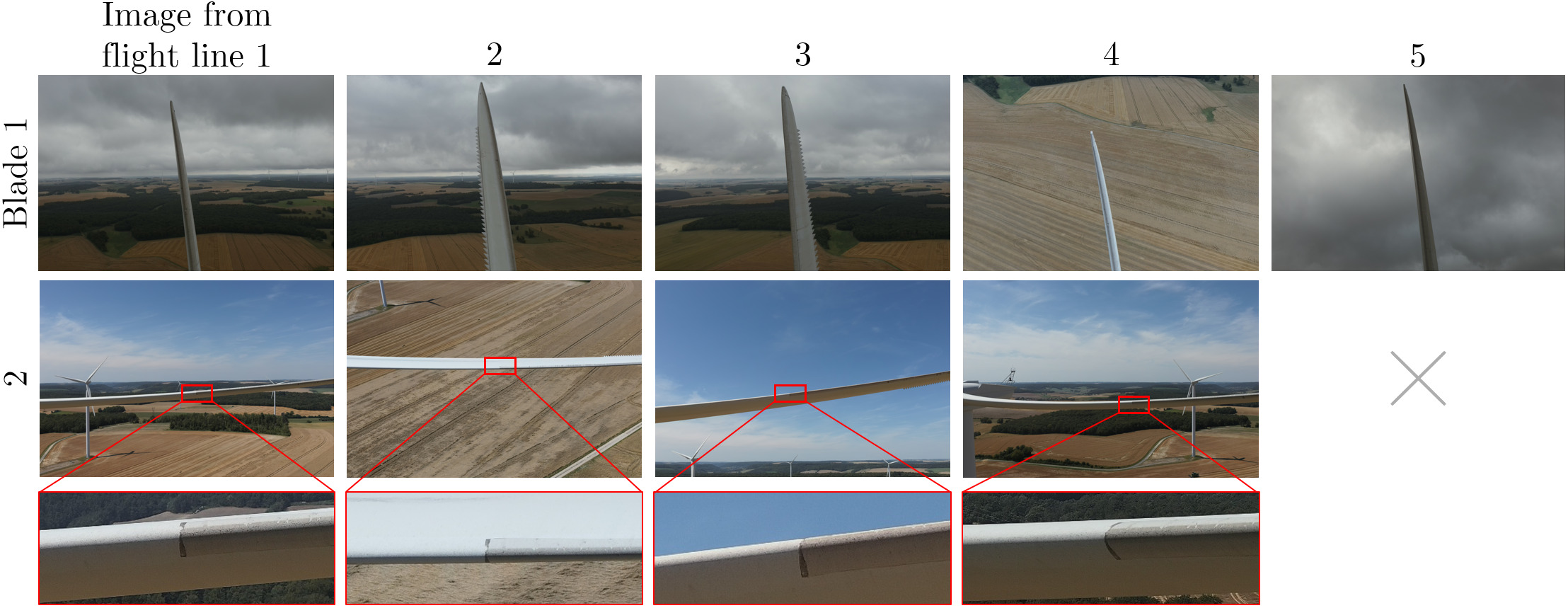}
	\caption{Representative images of both blades captured from each flight line.}
	\label{fig:image_samples}
\end{figure*}

Representative images from each flight line are shown in Fig. \ref{fig:image_samples}. Blade 1 (vertical, cloudy conditions) includes 104 images from two azimuthal and two polar angles, whereas Blade 2 (horizontal, sunny conditions) includes 121 images from two azimuthal and one polar angle.

\subsection{Preprocessing} \label{subsection:preprocessing}

\begin{figure*}[ht]
	\centering
	\includegraphics[width=0.85\linewidth]{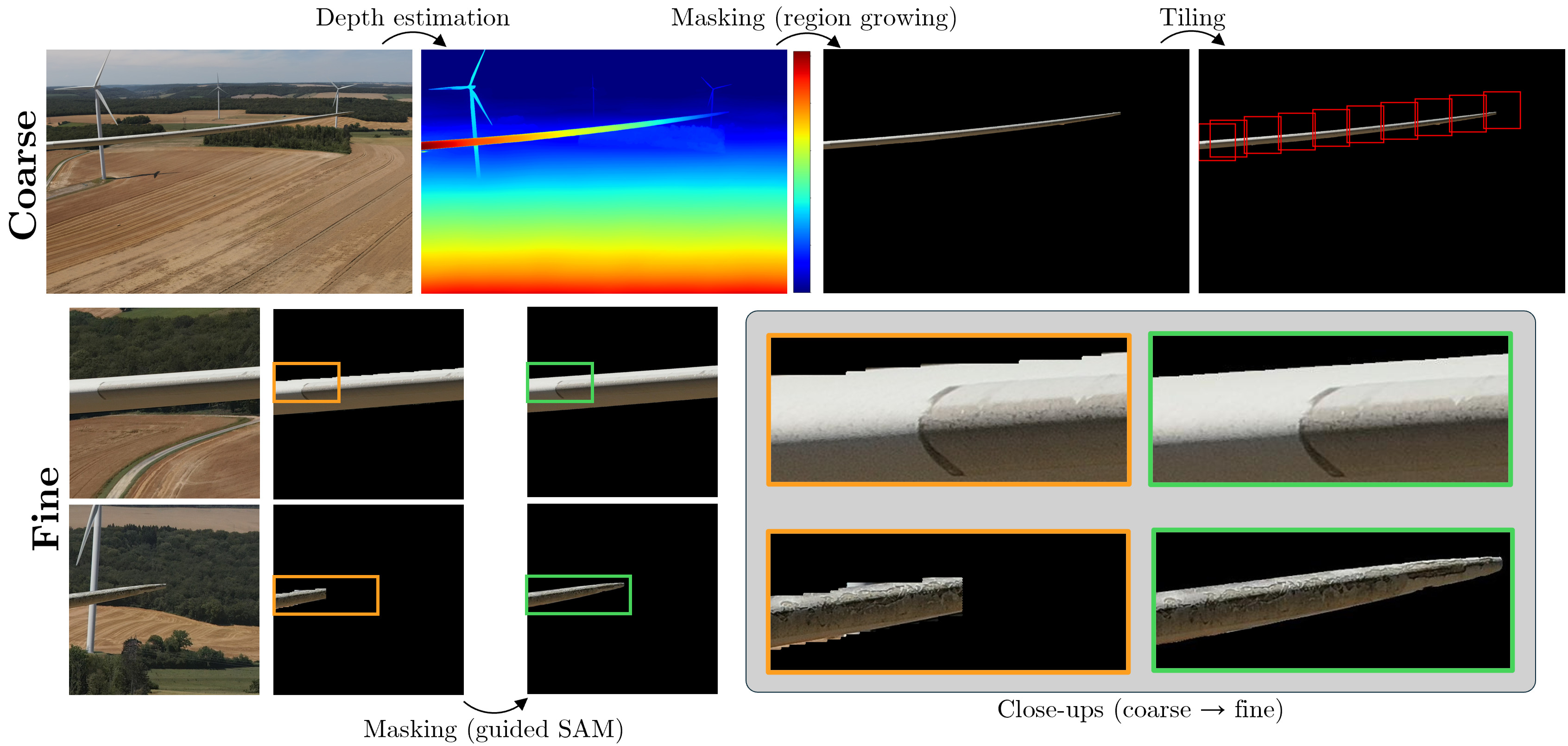}
	\caption{Preprocessing pipeline. The coarse stage (top row) estimates monocular depth from a downscaled image, generates a mask via seeded region growing, and computes tile coordinates (Alg. \ref{alg:tiling}). In the fine stage, tiles are processed at full resolution, and SAM refines the coarse masks.}
	\label{fig:preprocessing}
\end{figure*}

Sparse 3D reconstruction begins with keypoint detection and matching. To improve robustness, we apply two preprocessing steps: \textit{background masking} and \textit{image tiling} (Fig.~\ref{fig:preprocessing}).

\textbf{Background masking} produces a binary mask ($0$ = background, $1$ = foreground) to isolate the blade, mitigate dynamic backgrounds (e.g., clouds, nearby turbines), and constrain keypoint matching to foreground regions, improving correspondence quality.

\textbf{Image tiling} divides each image into patches to: (1) reduce computation by avoiding large background areas, and (2) preserve local detail, since deep keypoint detectors are trained on low-resolution inputs ($<$ 1 MP) and downscaling full images would remove fine blade structure.

We compute \textbf{coarse masks} using DepthAnythingV2 \cite{yang2024depth} on downscaled images. Unlike prior work that relies on a static depth threshold to separate foreground from background \cite{sterckx2025accurate}, we adopt a dynamic \textbf{seeded region-growing} strategy \cite{adams1994seeded}: starting from the closest pixel (the \emph{seed}), the mask expands while depth gradients remain below a threshold $d_\text{grad}^\text{max}$, improving segmentation when blade and background depths overlap (see Fig. \ref{fig:preprocessing}).

For \textbf{foreground-focused tiling}, the coarse mask (upscaled) guides extraction of the blade’s principal axis via PCA. Patch centers are sampled along this axis at stride $S$, tiles of size $P$ are extracted, and only patches overlapping the foreground are retained (Alg.~\ref{alg:tiling}).

\begin{algorithm}[!h]
	\caption{Foreground-Focused Image Tiling.} \label{alg:tiling}
	\KwData{Coarse mask $M$, patch size $P$, stride $S$}
	\KwResult{List of patch coordinates}
	
	Extract foreground points $C$ from $M$\;
	Compute centroid $\boldsymbol{\mu}$ and principal axis $\mathbf{v}$ using PCA\;
	Project $C$ onto $\mathbf{v}$ and get range $[p_{\min}, p_{\max}]$\;
	
	\For{$p_i \in [p_{\min}, p_{\max}]$ with step $S$}{
		Center $c_i = \boldsymbol{\mu} + p_i \mathbf{v}$\;
		Clamp patch at $c_i$ to image bounds\;
		\If{patch intersects $M$}{
			Add patch to output\;
		}
	}
	
	\Return patch coordinates\;
\end{algorithm}

Finally, masks are refined using \textbf{Segment Anything (SAM)} \cite{kirillov2023segment} on native-resolution tiles, using the coarse mask as a prompt to produce sharper \textit{fine masks}. Fig. \ref{fig:preprocessing} illustrates the full preprocessing pipeline.

\subsection{Sparse 3D Reconstruction}

Sparse 3D reconstruction estimates camera poses and a sparse point cloud, forming the basis for dense reconstruction. \citet{sterckx2025accurate} proposed a two-stage pipeline: tiles are matched at downscaled resolution to find coarse overlaps, then refined at full resolution. This fails for drone imagery at typical inspection distances, as downscaling removes critical detail. Brute-force matching (testing all tile pairs) is more robust but scales poorly ($\mathcal{O}(K^2 T^2)$, $K$ = number of images, $T$ = average tiles per image) and is infeasible for large datasets (taking $\approx$ 0.5 s per 1 MP image pair on an Nvidia 4070 Ti Super). We address this by leveraging GPS metadata and prior blade geometry to enable scalable matching.
\begin{figure*}[ht]
	\centering
	\includegraphics[width=0.65\linewidth]{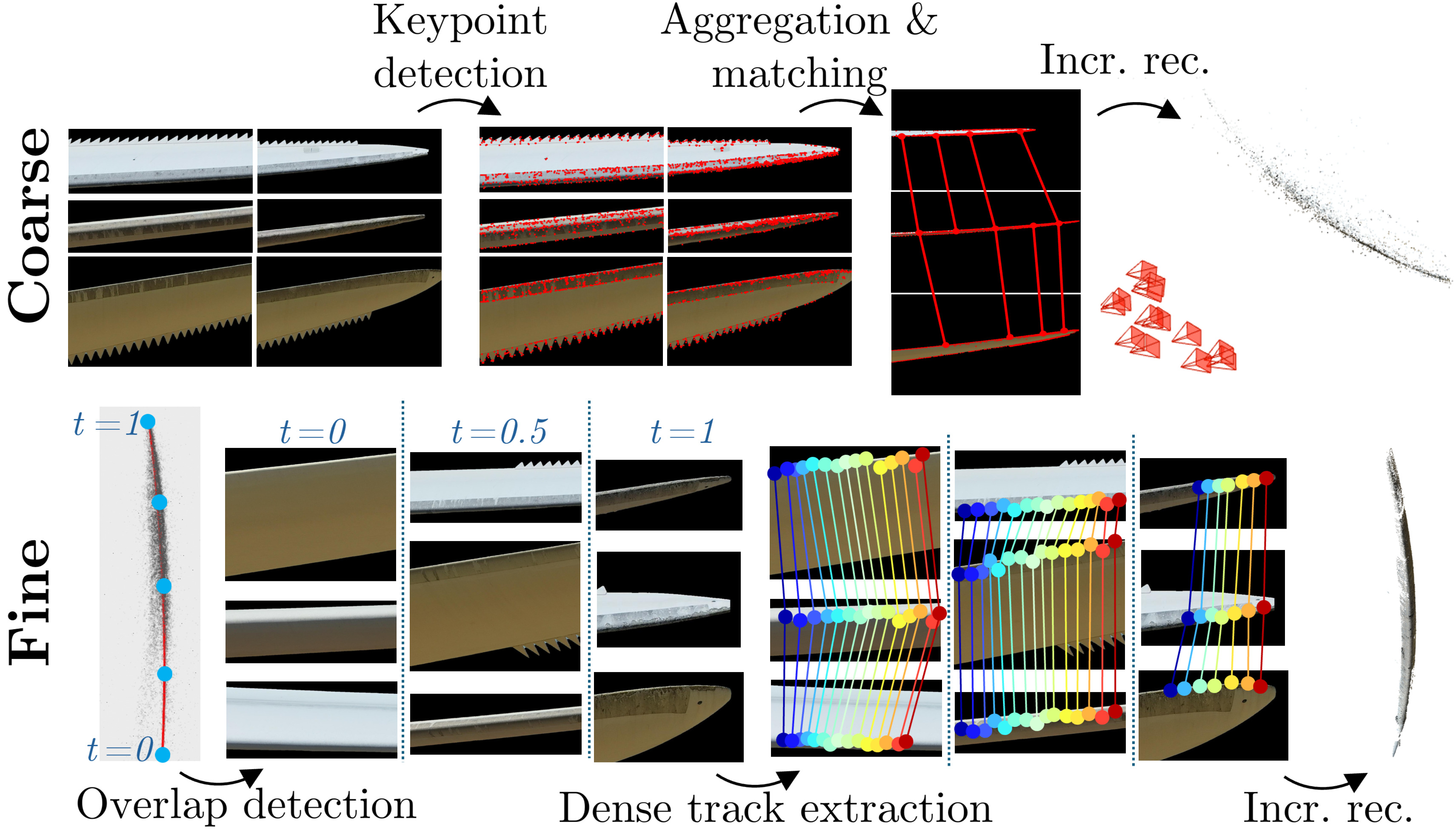}
	\caption{Coarse-to-fine sparse reconstruction pipeline. Coarse stage: keypoints are detected in tiles, aggregated, and sparsely matched; matches and GPS/IMU poses (not shown) are fed into COLMAP. Fine stage: the leading-edge curve is estimated to locate overlapping patches; dense matches are computed, aggregated, and again input to COLMAP.}
	\label{fig:sparse_pipeline}
\end{figure*}
\\[0.1cm]
\noindent \textbf{GPS-guided coarse reconstruction.}  
We detect sparse keypoints using SuperPoint \cite{detone2018superpoint} in each foreground tile, aggregate them into full-image keypoints, and match across images with LightGlue \cite{lindenberger2023lightglue}. The resulting correspondences, combined with UAV GPS/IMU data, are fed into COLMAP, which initializes camera poses, filters matches, and performs incremental reconstruction \cite{colmap}.  
The resulting coarse reconstruction, though noisy, is sufficiently accurate to improve upon GPS poses, capture the overall blade geometry, and initialize the next stage (Fig.~\ref{fig:sparse_pipeline}, top).
\\[0.1cm]
\noindent \textbf{Shape-guided fine reconstruction.}  
From the coarse point cloud, we extract the leading edge by fitting a second-order polynomial 3D curve $\mathbf{c}(t)$ to the sparse points. The fit uses RANSAC to robustly estimate the coefficients and yields a continuous parametric space curve that represents the blade's leading edge.

We uniformly sample $L$ points along $\mathbf{c}(t)$ and project them into all images using coarse poses. 
Each 3D point is projected into all images, and if it lies within the image bounds, a patch around the point is extracted. 
From overlapping patches, random subsets of size $n$ are sampled until all patches are included.
The computational cost of matching now scales as $\mathcal{O}(LK')$, where $K'$ is the mean number of images in which each space curve sample is observed.
Dense keypoint tracks are then generated from these subsets using the RoMa matcher \cite{edstedt2024roma}, following the approach of \cite{sterckx2025accurate}. These are fed into COLMAP for the final high-precision sparse reconstruction. 
Fig. \ref{fig:sparse_pipeline} (bottom) shows the fine reconstruction stage.

Before dense reconstruction, all images are undistorted using lens parameters estimated during bundle adjustment.

\subsection{Dense 3D Reconstruction}

We propose a custom Gaussian Splatting (GS) framework based on 2DGS \cite{2dgs}, which employs flat rather than volumetric Gaussians for enhanced surface reconstruction, and supports rendering of RGB, depth, and opacity. Beyond visual losses, the \textit{normal consistency} loss aligns disk normals with depth-map normals to ensure splats approximate true surfaces, while the \textit{depth distortion} loss aligns non-occluded disks along each ray, reducing ambiguity from spatially separated but visually similar splats.

To adapt 2DGS for high-resolution UAV imagery of turbine blades, we introduce:
\begin{enumerate}
	\item \textbf{Tiled training and rendering} to process high-resolution images and retain fine detail.
	\item \textbf{Soft background masking} to isolate the blade while tolerating imperfect boundaries.
	\item \textbf{Blade-specific densification} to prioritize detail at the leading edge.
\end{enumerate}
In addition, we integrate \textbf{exposure normalization} and \textbf{motion/defocus deblurring} to handle lighting variations and blur, combining techniques from prior work.

\subsubsection{Soft Background Masking}
Object-centric reconstruction (foreground isolation) is achieved by penalizing non-transparent background renderings via
\begin{equation}\label{eq:mask_loss}
	\mathcal{L}_\text{mask} = \lambda_\text{mask} \sum_{\mathbf{p}} (1 - M(\mathbf{p})) \cdot (O(\mathbf{p}))^2,
\end{equation}
where $M(\mathbf{p})$ and $O(\mathbf{p})$ are the binary mask value and rendered opacity at pixel $\mathbf{p}$, and $\lambda_\text{mask}$ is a hyperparameter. Since transparent Gaussians are removed during optimization \cite{kerbl20233d}, this effectively eliminates primitives outside the blade.

\begin{figure}[ht]
	\centering
	\includegraphics[width=0.99\linewidth]{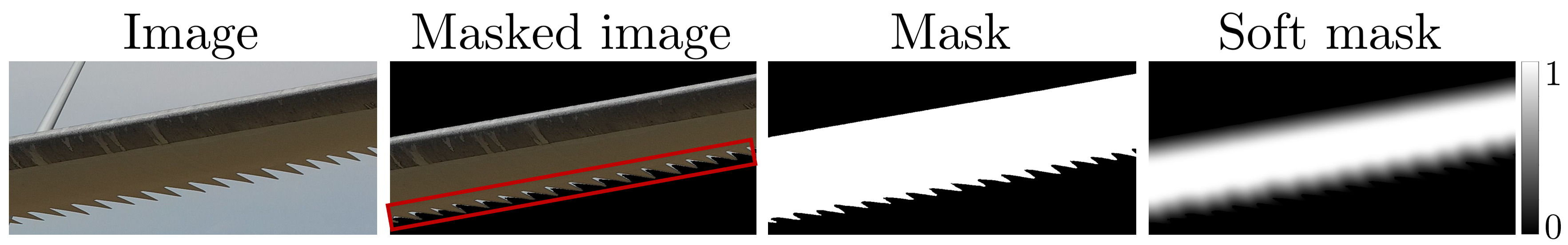}
	\caption{Example of soft masking to mitigate artifacts from imperfect segmentation.}
	\label{fig:soft_masking}
\end{figure}

Unlike prior work \cite{rogge2025object}, we mitigate the impact of potential mask imperfections by using \textbf{soft masks}, in which pixels near the boundaries take continuous values between 0 and 1. This mask is obtained by Gaussian smoothing the binary mask with a standard deviation $\sigma_\text{soft}$ (see Fig.~\ref{fig:soft_masking}).

All other loss terms are similarly weighted by the soft mask $M(\mathbf{p})$, improving robustness to segmentation noise and enhancing reconstruction near boundaries.

\subsubsection{Tiled Full-Resolution Processing}

To process 45 MP images without downscaling, which would otherwise degrade fine surface details, we adopt a tiled training and rendering strategy. Instead of feeding full images, which quickly exceed GPU memory limits, we further crop the preprocessed tiles (Section~\ref{subsection:preprocessing}) to the foreground bounding box with a 5\% margin, and use these as independent training inputs. This reduces both rendering time and memory usage.

Our tiled GS pipeline requires four key modifications: (1) computing virtual \textit{camera parameters} for tiles, (2) adapting the \textit{image scheduler}, (3) aggregating \textit{tile-based losses}, and (4) \textit{sharing camera-specific modules} across tiles.
\\[0.1cm]
\noindent \textbf{Camera parameters of tiles.}
Tile rendering requires virtual intrinsics adjusted for cropping. While the camera pose and focal length remain unchanged, the principal point shifts by $(-x_0, -y_0)$ for a tile with top-left corner $(x_0, y_0)$. This modifies the intrinsic and projection matrices, and necessitates an extension of the GS renderer to support off-axis optical centers.
\\[0.1cm]
\noindent \textbf{Tiled training scheduler.}
In standard GS, training images are accessed in a randomized order. For tiled inputs, all tiles of a single image are processed consecutively, enabling per-tile losses to be combined into a consistent full-image loss. This batching-like strategy stabilizes optimization without increasing memory overhead.
\\[0.1cm]
\noindent \textbf{Aggregating tile-based losses.}
Per-pixel losses (e.g., $\mathcal{L}_1$, $\mathcal{L}_\text{SSIM}$) are weighted by the product of the soft mask and an \textit{overlap map} $W(\mathbf{p})$, which encodes how many tiles cover each pixel to avoid overcounting in overlapping tiles. This weighting ensures correct contributions from visible foreground areas. The total image loss is then computed as the weighted mean across tiles, preserving equivalence with the full-image loss in the non-tiled setup.
\\[0.1cm]
\noindent \textbf{Sharing camera-specific modules.}
All tiles of an image share the same capture conditions (lens, motion, and exposure). To maintain consistency across tiles, we share a single instance of the exposure normalization module (see Section \ref{sec:capture_aware_rendering}) across all tiles of an image. When modeling other image-specific parameters, such as camera motion or depth of field, they should likewise be shared across tiles.

\subsubsection{Adaptive Densification}

\begin{figure}[ht]
	\centering
	\includegraphics[width=0.85\linewidth]{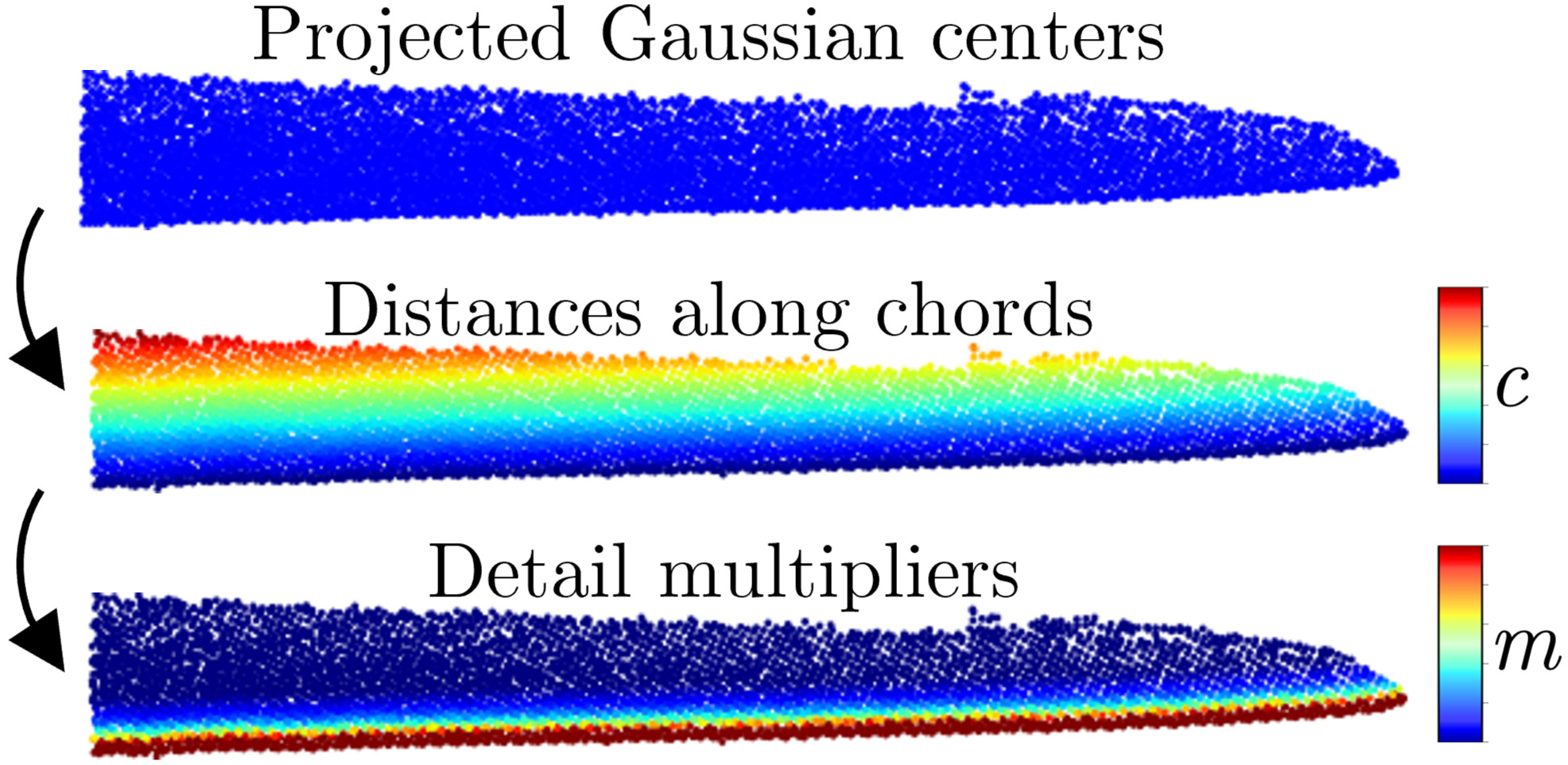}
	\caption{Computation of detail multipliers from Gaussian positions.}
	\label{fig:ada_densif}
\end{figure}

To capture fine-scale geometry in critical regions, we introduce adaptive Gaussian densification that prioritizes the leading edge, an area prone to erosion. By allocating more Gaussians here, we increase detail without uniformly raising the total Gaussian count, balancing fidelity and efficiency.

During each training iteration where high-loss Gaussians are cloned or split, we identify leading-edge Gaussians by projecting their centers onto the blade’s principal axes (spanwise and chordwise). The blade is sliced along the spanwise axis, and in each slice, the point farthest along the chordwise axis defines the leading-edge reference.

For each Gaussian in a slice, its chordwise distance $c$ from the reference determines a detail multiplier $m$:
\begin{equation}
	m(c) = \max\!\left( 1, \, \min\!\left( \frac{\tau}{c}, \, m_{\text{max}} \right) \right),
\end{equation}
where $m_{\text{max}}$ is the maximum multiplier value and $\tau$ defines the chordwise range over which higher detail is applied. The densification threshold for each Gaussian is then scaled by $1/m$, so Gaussians nearer the leading edge densify more readily.

Fig.~\ref{fig:ada_densif} exemplifies this process.

\subsubsection{Capture-Aware Rendering}\label{sec:capture_aware_rendering}

\begin{figure*}[ht]
	\centering
	\includegraphics[width=0.75\linewidth]{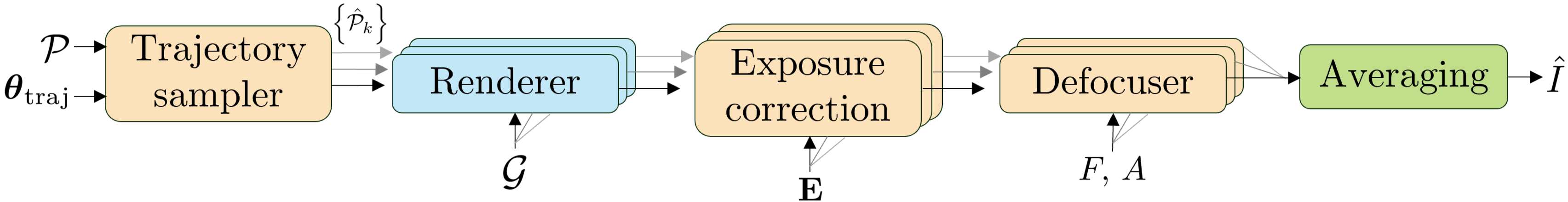}
	\caption{Camera motion during exposure is emulated by sampling poses $\mathcal{P}_k$ along a trajectory around input pose $\mathcal{P}$, parameterized by learnable $\boldsymbol{\theta}_\text{traj}$ \cite{bad_gaussians}. Each snapshot of the Gaussian model ($\mathcal{G}$) is color-corrected using learnable exposure parameters $\mathbf{E}$ \cite{kerbl2024hierarchical} and defocused using learnable focal distance $F$, aperture $A$, and depth \cite{shen2025dof}, then averaged to produce the final motion-blurred render $\hat{I}$.}
	\label{fig:capture_aware_rendering}
\end{figure*}

To account for real-world image degradations such as exposure variation, motion blur and defocus blur, our framework allows these effects to be modeled as part of the \textbf{rendering process} rather than as scene properties. This decoupling helps ensure that perceptual degradations are not mistakenly attributed to the 3D model.

We integrated modules for motion and defocus deblurring \cite{bad_gaussians} \cite{shen2025dof}, and per-image exposure normalization \cite{kerbl2024hierarchical}, enabling a fully differentiable, capture-aware rendering pipeline (Fig.~\ref{fig:capture_aware_rendering}). 

\begin{figure}[ht]
	\centering
	\includegraphics[width=0.9\linewidth]{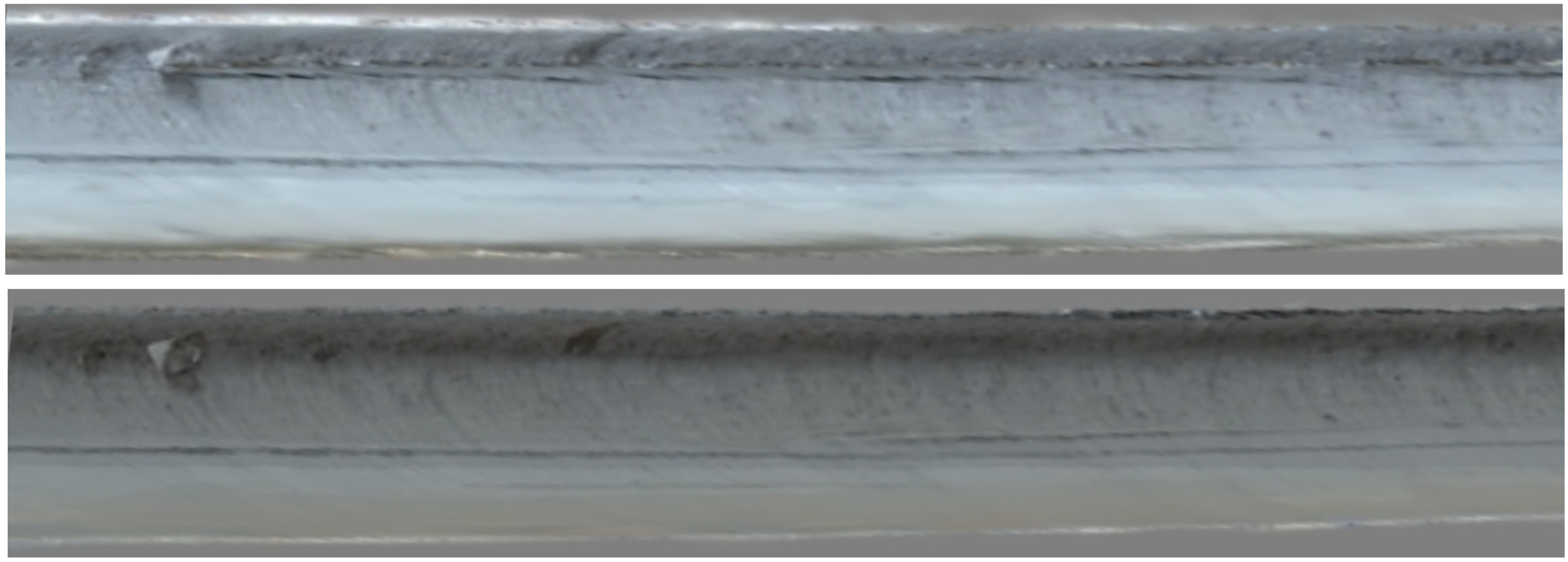}
	\caption{Models rendered from unseen poses. Top/bottom: trained without/with exposure normalization.}
	\label{fig:exposure_comp}
\end{figure}

Exposure normalization proved essential to mitigate artifacts (Fig. \ref{fig:exposure_comp}) due to appearance changes across viewing angles (Fig. \ref{fig:image_samples}). Motion and defocus blur were negligible in our imagery, and enabling the corresponding modules had minimal effect on reconstruction quality. We mention them for completeness, as we expect their impact to be greater on datasets with stronger degradations, but omit them from further discussion.

\subsection{Hyperparameters}

In our experiments, we use the following parameters:
$d_\text{grad}^\text{max}=0.1$, 
$P=(1024,1024)$, 
$L=20$, 
$n=5$, 
$ \lambda_\text{mask} = 100 $,
$ \sigma_\text{soft} = 5$, 
$ m_\text{max} = 5 $, and
$ \tau = 0.2\bar{c} $ (where $ \bar{c} $ is the chord length of the mid-span blade slice).
All other dense reconstruction parameters follow the default 2DGS settings.


\section{RESULTS}

We evaluate our proposed pipeline against alternative 3D reconstruction software. In the absence of ground truth 3D models, our evaluation relies on a combination of quantitative metrics for the sparse reconstruction and qualitative analysis for the dense reconstruction, including a detailed ablation study of our custom GS framework.

\subsection{Sparse Reconstruction}

\begin{figure*}[ht]
\centering
\begin{minipage}{0.34\linewidth}
    \centering
    \includegraphics[width=\linewidth]{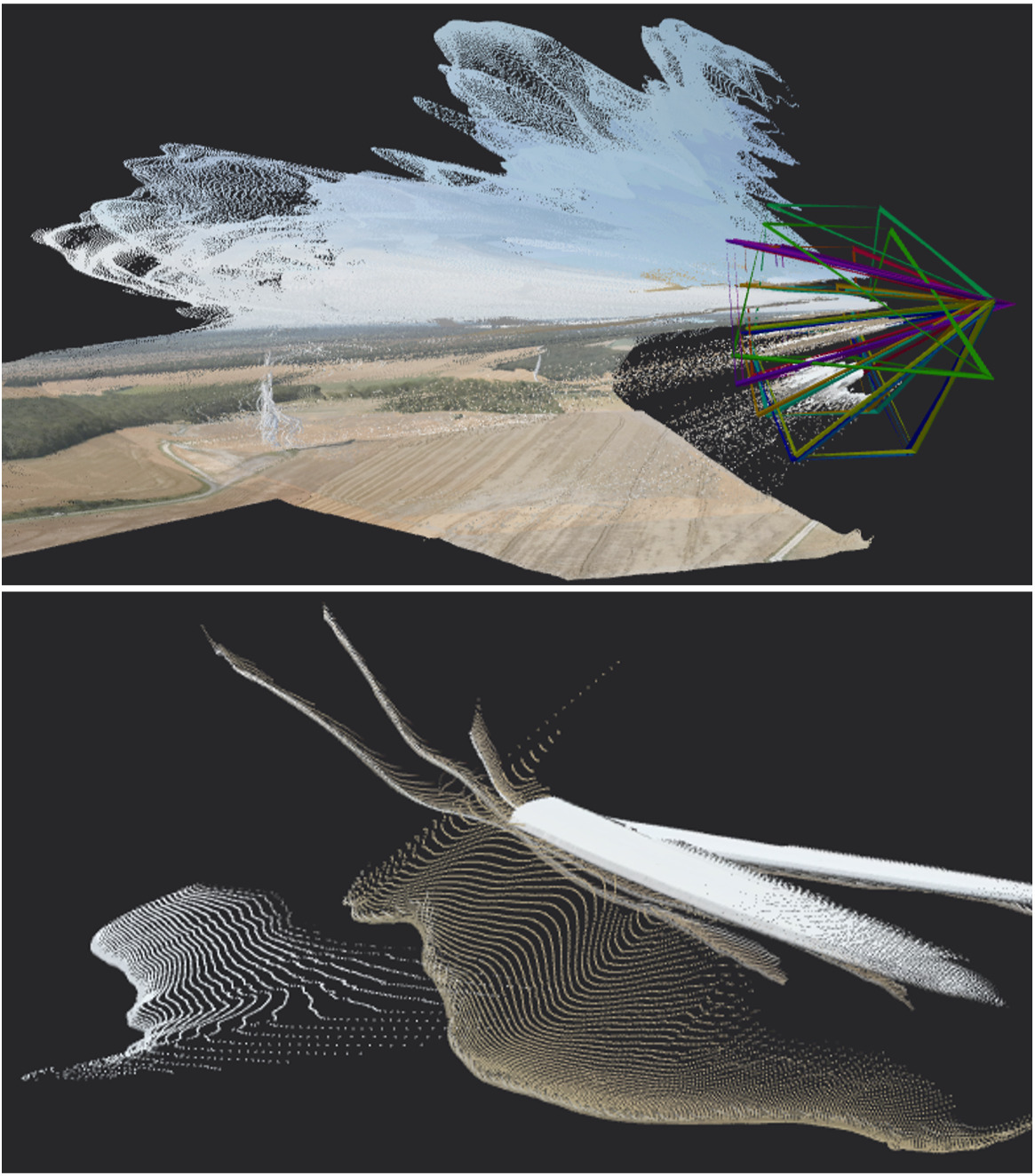}
    \captionof{figure}{VGGT reconstruction on downscaled Blade 2 images. Top/bottom: 3D reconstructions from unmasked/masked inputs. Masking reduces noisy sky points but causes duplicated, misaligned blade surfaces.}
    \label{fig:vggt_results}
\end{minipage}
\hfill
\begin{minipage}{0.65\linewidth}
    \centering
    \includegraphics[width=\linewidth]{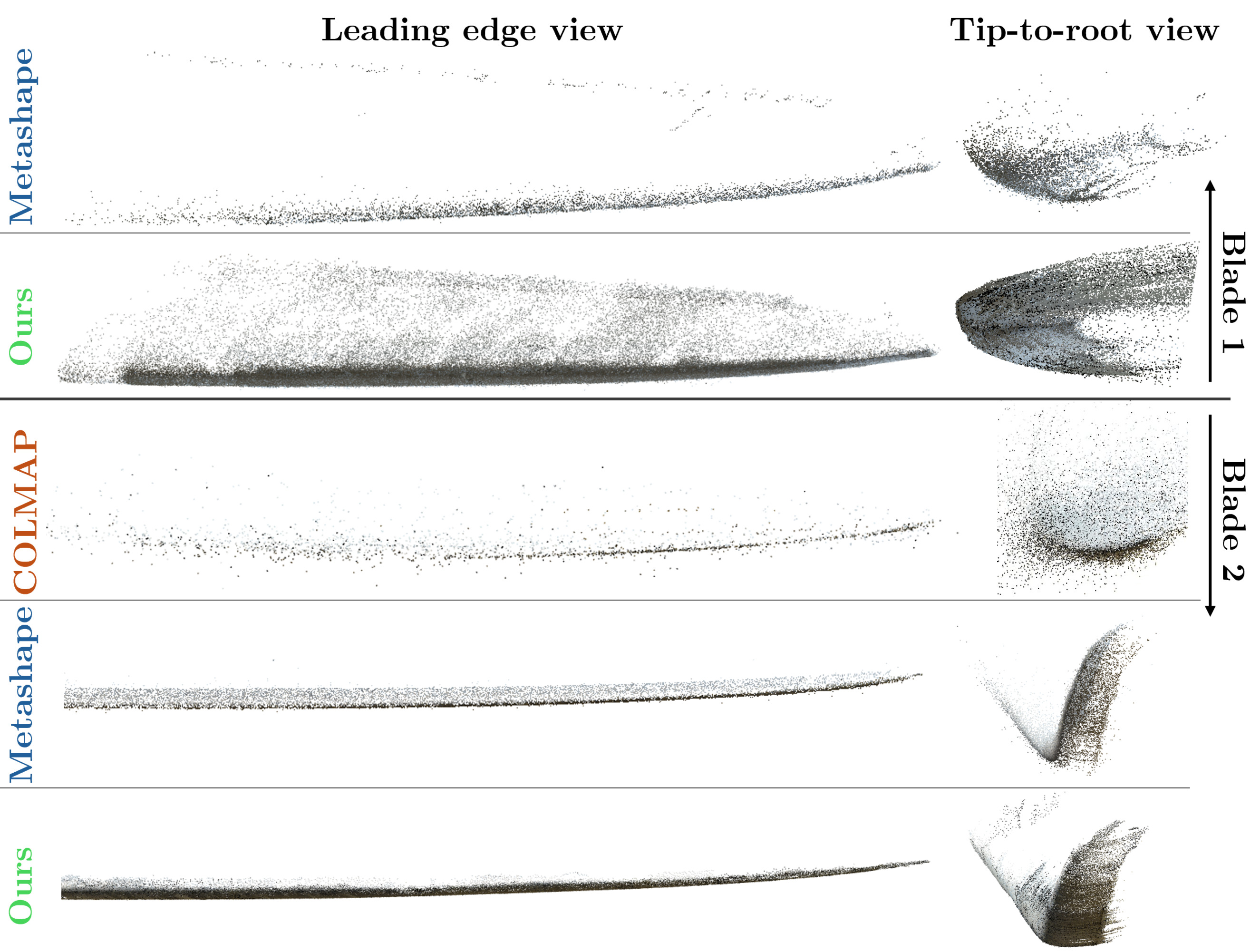}
    \captionof{figure}{Visual comparison of sparse 3D reconstructions.}
    \label{fig:sparse_models_combined}
\end{minipage}
\end{figure*}

We evaluate the quality of sparse reconstructions produced by our pipeline against two widely used baselines: the leading open-source framework, COLMAP \cite{colmap}, and the leading commercial software, Metashape \cite{metashape}. For a fair comparison, both tools were provided with the GPS pose priors and our background masks, and their reconstruction quality settings were set to the maximum. These methods rely on classical feature matching, which is well suited to high-resolution imagery. Recent deep learning-based methods, including Detector-Free SfM \cite{detectorfreesfm}, MASt3R \cite{leroy2024grounding}, and VGGT \cite{wang2025vggt}, rely on deep feature extractors limited by GPU memory and are typically trained and used on inputs of $\sim$1 MP or less. Applied to our full-resolution data, all three quickly hit memory bottlenecks. Downscaling to the model's expected training resolution drastically reduces detail, preventing successful reconstruction. Fig. \ref{fig:vggt_results} illustrates such a failure using VGGT, a state-of-the-art sparse reconstruction method.

We evaluate sparse reconstruction quality, which impacts the subsequent dense reconstruction, using the following metrics:
\begin{enumerate}
	\item \textbf{RMS Reprojection Error} ($E$): Measures geometric consistency; lower values indicate better alignment of 3D points with image observations.
	\item \textbf{Point count} ($N$): Total number of reconstructed 3D points. A higher count strengthens initialization for dense reconstruction and, when coupled with low $E$, indicates a reliable model \cite{multiple_view_geometry}.
	\item \textbf{Mean track length} ($T$): Average number of images used to triangulate each 3D point. Longer tracks imply more robust correspondences.
\end{enumerate}

Table \ref{tab:sparse_quality} reports these metrics for both wind turbine blades. To visually evaluate completeness, smoothness, and noise, we render the corresponding sparse models in Fig. \ref{fig:sparse_models_combined}.

\begin{table}[ht]
	\caption{Comparison of sparse reconstruction metrics across pipelines for the two blades.}
	\label{tab:sparse_quality}
	\centering
	\setlength{\tabcolsep}{2.5pt} 
	\renewcommand{\arraystretch}{1.1} 
	\begin{tabular}{|l||c|c|c||c|c|c|}
		\hline
		\multicolumn{1}{|l||}{} & \multicolumn{3}{c||}{\textbf{Blade 1}} & \multicolumn{3}{c|}{\textbf{Blade 2}} \\
		\cline{2-7} 
		\multicolumn{1}{|l||}{} & $E\!\downarrow$ & $N\!\uparrow$ & $T\!\uparrow$ & $E\!\downarrow$ & $N\!\uparrow$ & $T\!\uparrow$ \\
		\hline
		COLMAP     & ---   & ---    & ---  & 1.50  & 22165  & 2.95 \\
		Metashape  & 0.42  & 6781   & 2.57 & 0.41  & 23177  & 3.92 \\
		Ours       & 0.25  & 427376 & 3.67 & 0.18  & 223348 & 4.73 \\
		\hline
	\end{tabular}
\end{table}

\subsection{Dense Reconstruction}

This section presents the dense reconstruction results, compares them with alternative methods, and assesses the impact of our novel extensions to 2DGS.

We first compare our pipeline with COLMAP's MVS \cite{schonberger2016pixelwise} and Metashape. For a fair comparison, all methods were provided with the same camera poses (those of our sparse reconstruction). Deep learning-based MVS alternatives were not evaluated, as they are trained on and restricted to low-resolution imagery, and run into memory bottlenecks at higher resolutions. 

\begin{figure*}[!t]
	\centering
	\begin{subfigure}[b]{0.49\linewidth}
		\centering
		\includegraphics[width=\linewidth]{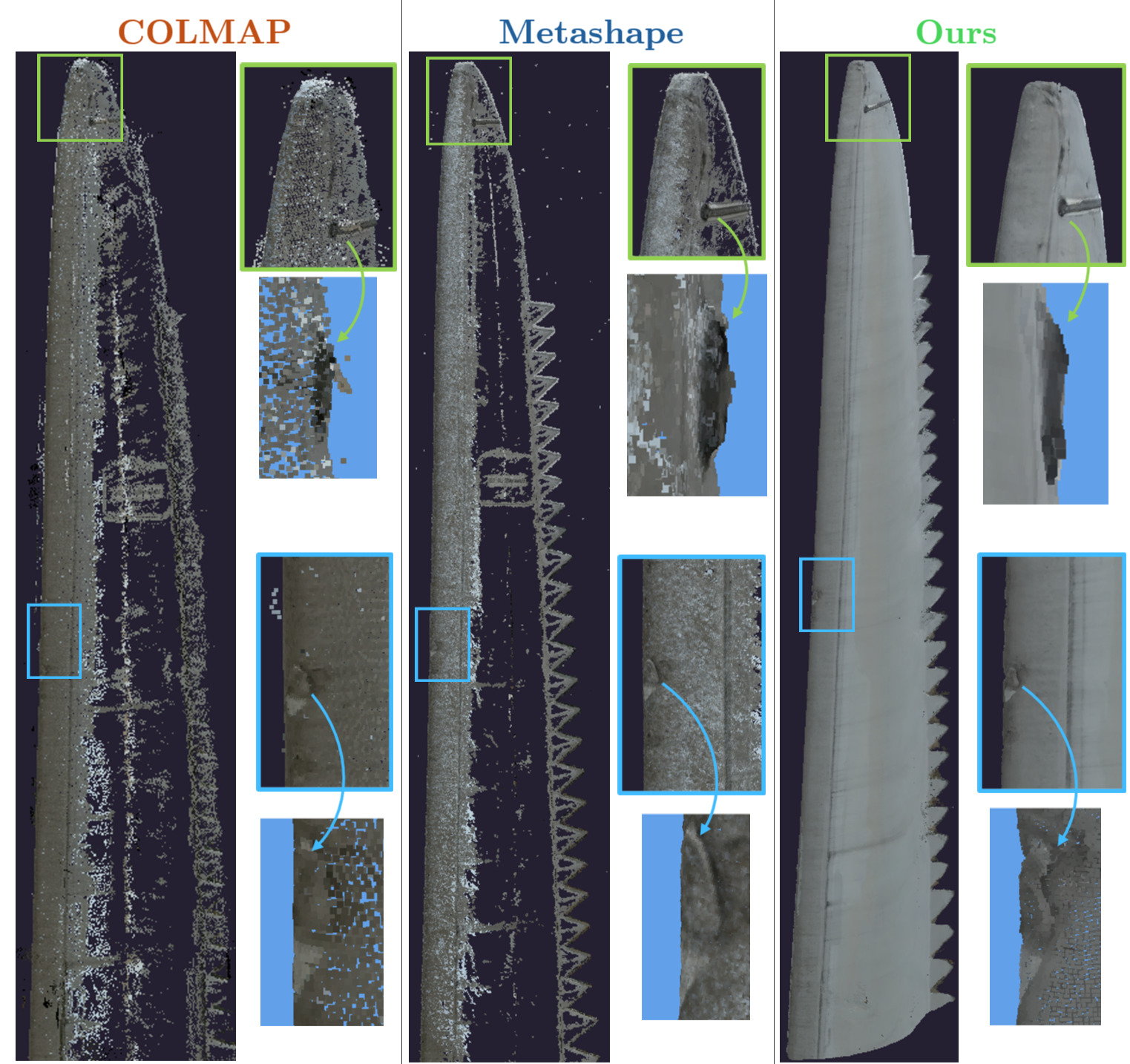}
		\caption{Dense reconstruction of Blade 1.}
		\label{fig:dense_blade1}
	\end{subfigure}
	\hfill
	\begin{subfigure}[b]{0.49\linewidth}
		\centering
		\includegraphics[width=\linewidth]{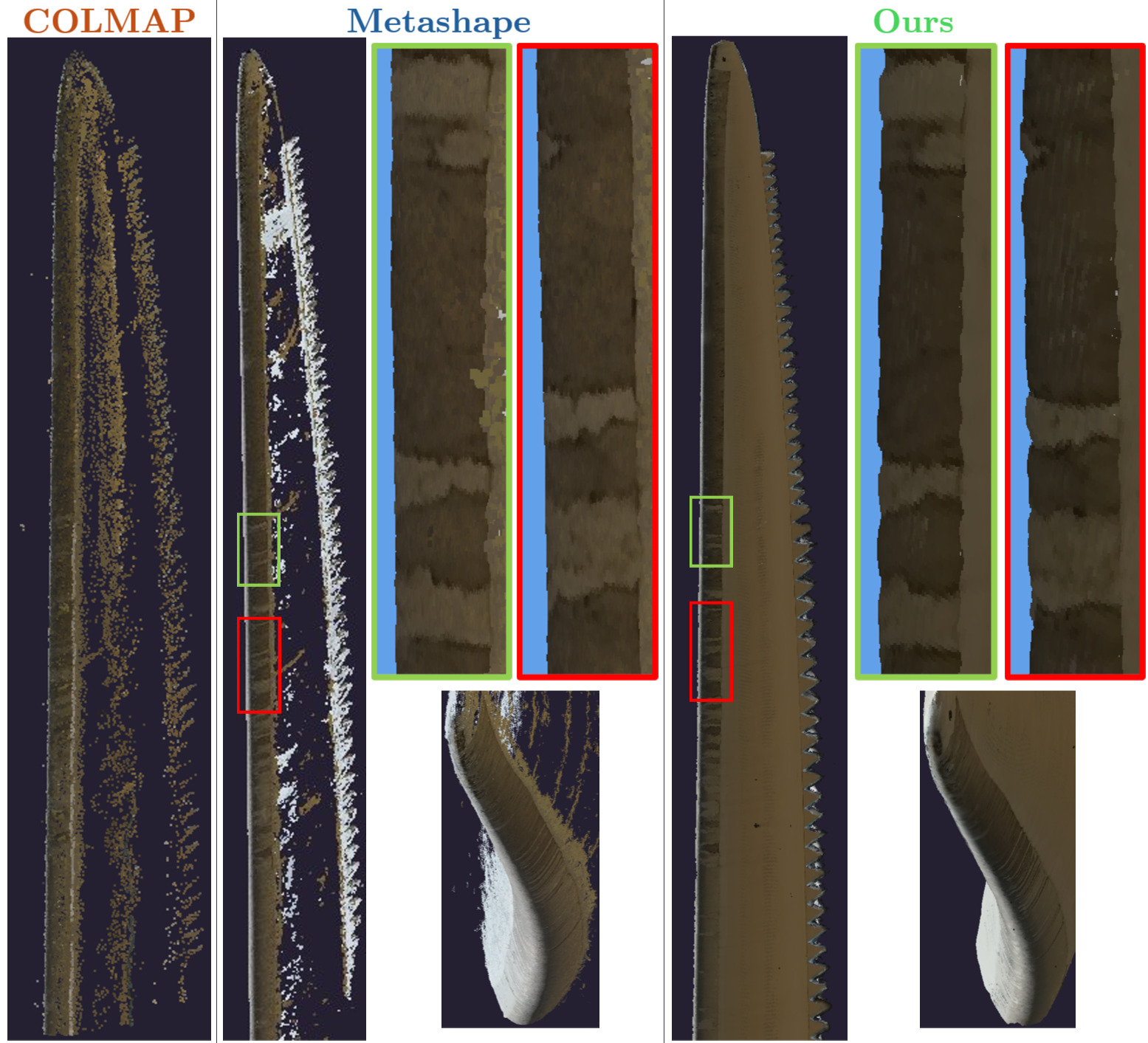}
		\caption{Dense reconstruction of Blade 2.}
		\label{fig:dense_blade2}
	\end{subfigure}
	\caption{Visual comparison of the dense 3D reconstructions for both blades.}
	\label{fig:dense_model_comparison}
\end{figure*}

In the absence of ground truth 3D data, our evaluation is qualitative, based on visual inspection, and focuses on completeness and surface detail. Fig. \ref{fig:dense_model_comparison} shows side-by-side comparisons of the dense 3D models of each blade produced by each method. For illustration, we converted our method's output into a dense point cloud by fusing the rendered depth maps of each image using Truncated Signed Distance Field (TSDF) fusion.

\begin{figure*}[!t]
	\centering
	\begin{subfigure}[b]{0.99\linewidth}
		\centering
		\includegraphics[width=\linewidth]{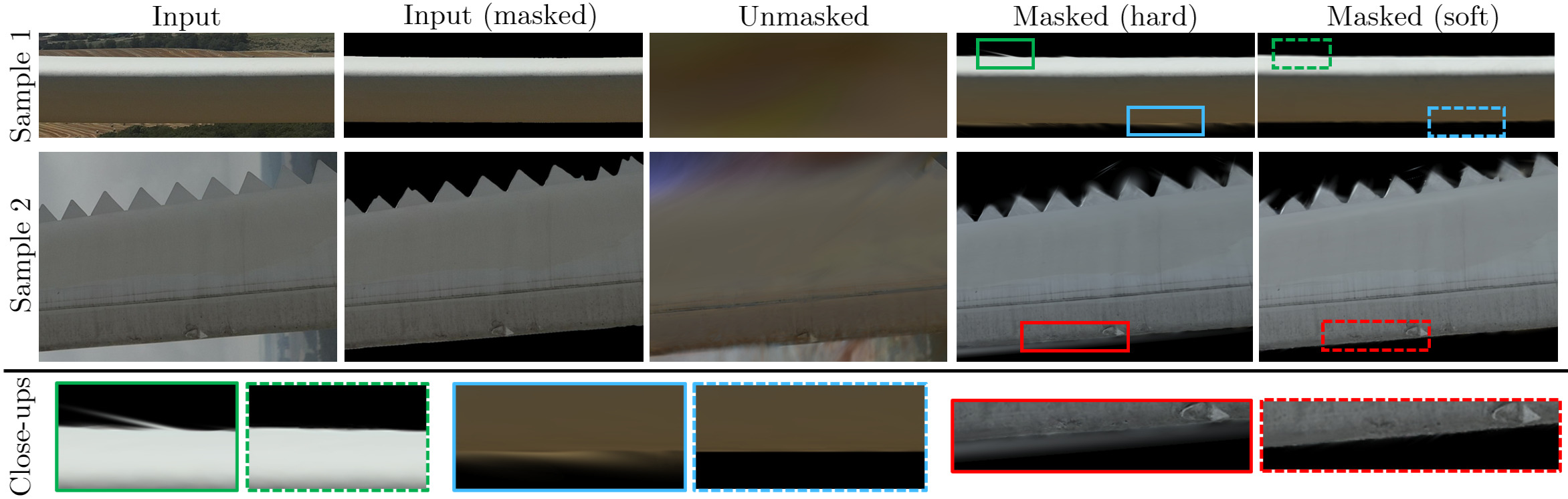}
		\caption{Effect of masking.}
		\label{fig:effect_masking}
	\end{subfigure}
	\\[1em] 
	\begin{minipage}[b]{0.48\linewidth}
		\centering
		\begin{subfigure}[b]{\linewidth}
			\centering
			\includegraphics[width=\linewidth]{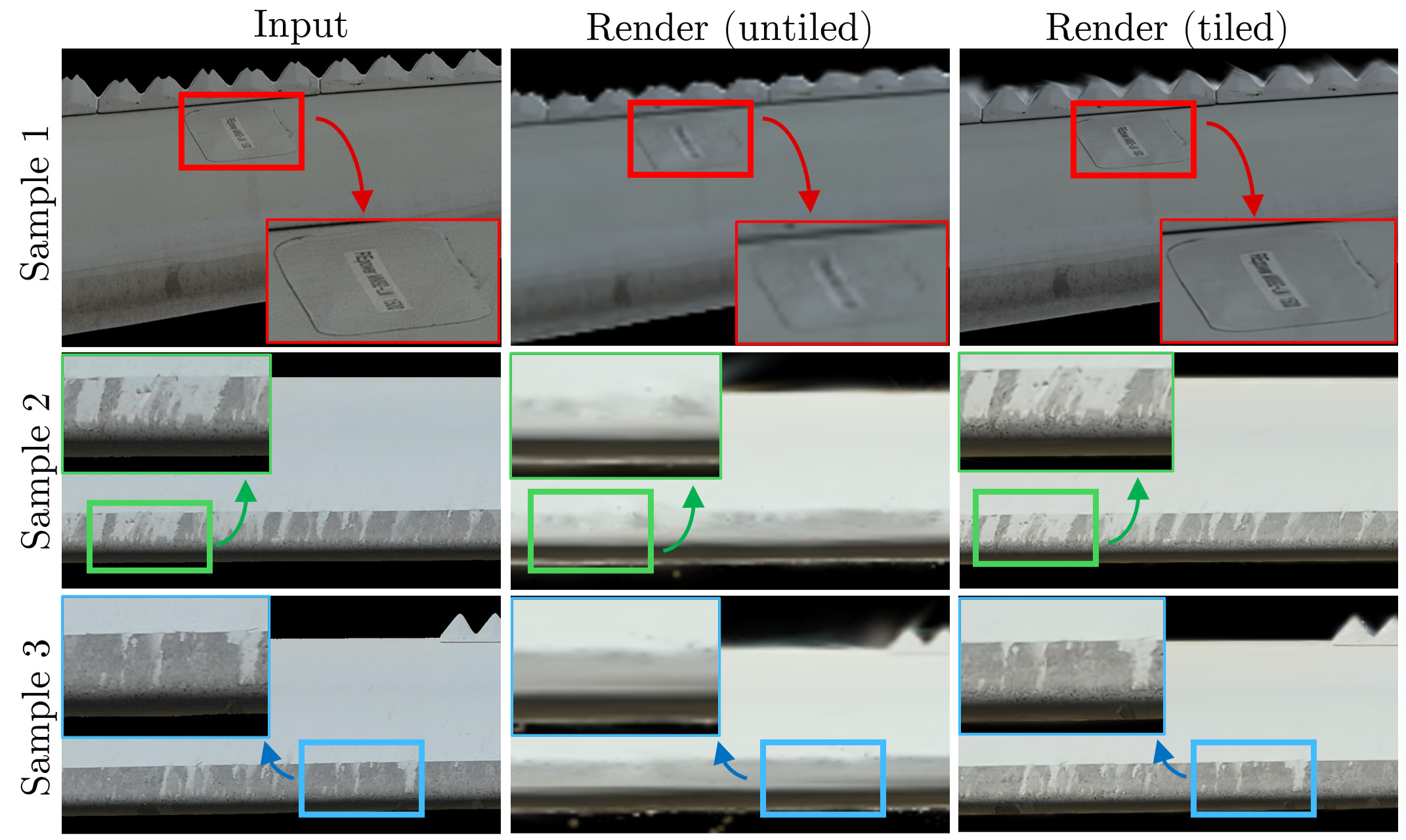}
			\caption{Effect of tiled training.}
			\label{fig:effect_tiling}
		\end{subfigure}
	\end{minipage}
	\hfill
	\begin{minipage}[b]{0.48\linewidth}
		\centering
		\begin{subfigure}[b]{\linewidth}
			\centering
			\includegraphics[width=\linewidth]{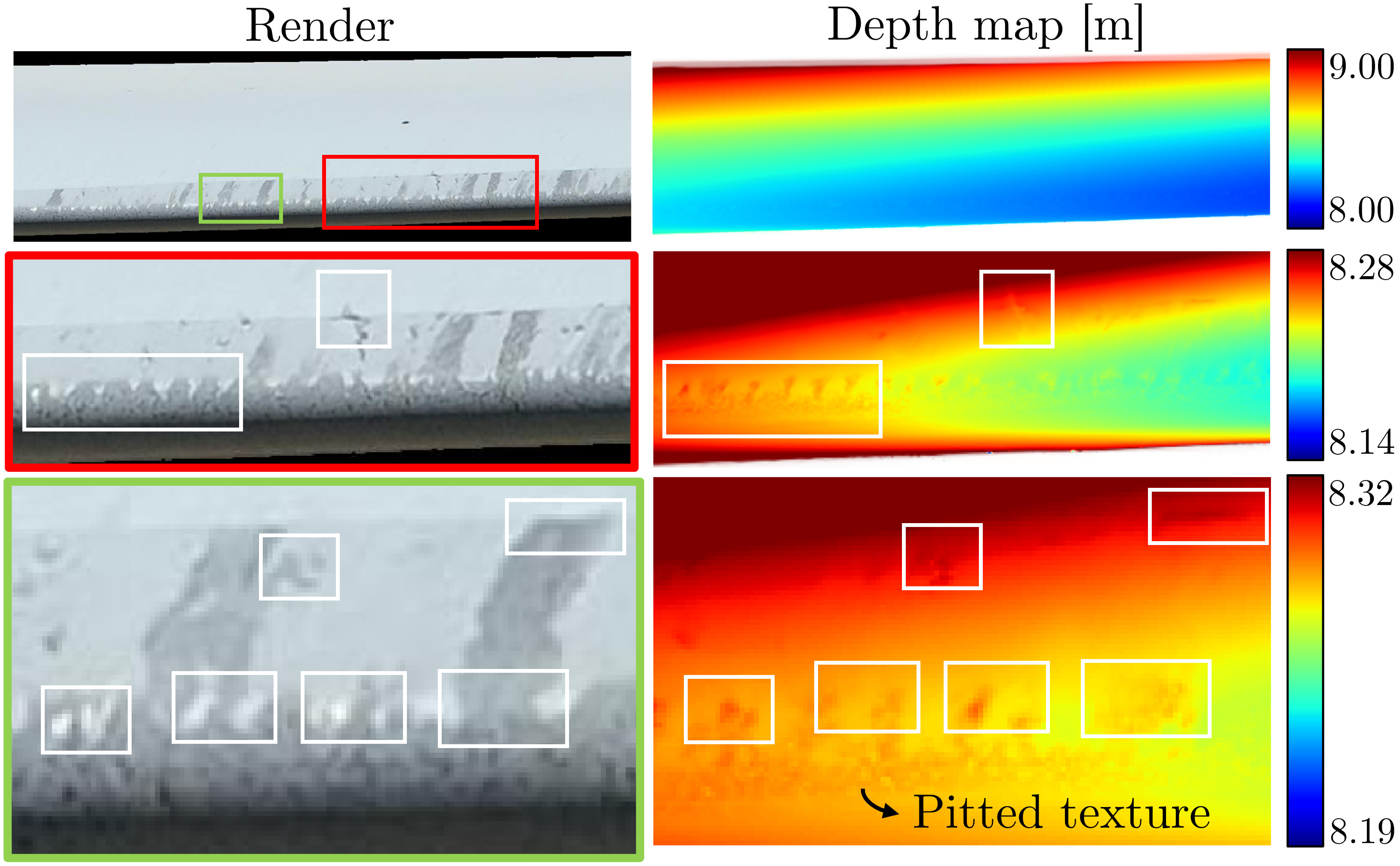}
			\caption{Example of achieved surface detail (all extensions enabled).}
			\label{fig:surface_detail}
		\end{subfigure}
	\end{minipage}
	\\[1em] 
	\begin{subfigure}[b]{0.99\linewidth}
		\centering
		\includegraphics[width=\linewidth]{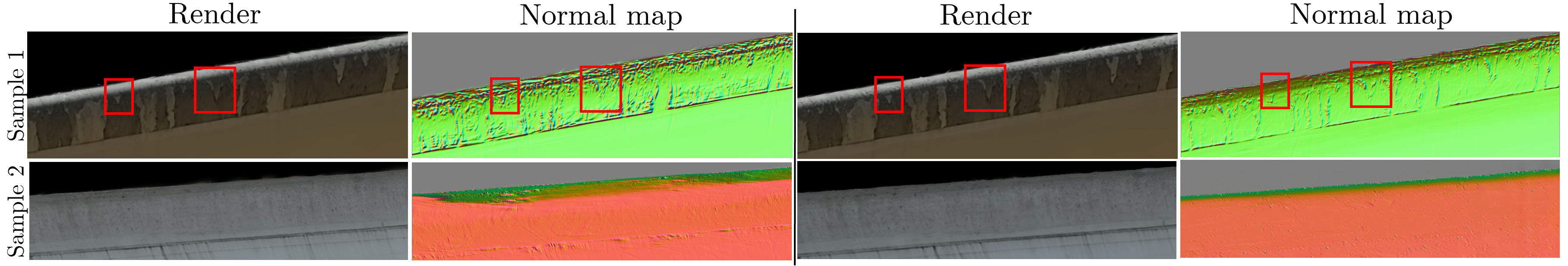}
		\caption{Effect of adaptive densification. Left/right: without/with adaptive densification.}
		\label{fig:effect_ada_densif}
	\end{subfigure}
	\caption{Illustrative examples demonstrating the impact of each proposed extension and achieved surface detail.}
	\label{fig:dense_ablation}
\end{figure*}

Next, we illustrate the impact of each extension to the 2DGS framework. Figure~\ref{fig:effect_masking} shows the effect of background masking, including a comparison of soft and hard masking. Figure~\ref{fig:effect_tiling} compares model detail when training on full images at $1/8$ resolution (the maximum size manageable with 16\,GB VRAM) against tiled training at full resolution. Figure~\ref{fig:effect_ada_densif} demonstrates the impact of adaptive densification on model appearance and geometry, visualized via color-coded normal maps. Finally, Figure~\ref{fig:surface_detail} exemplifies the achieved surface detail when enabling all extensions, illustrated by depth maps.

Finally, our tiling approach yielded an average tile size of 0.49\,MP and 9.1\,tiles per image, reducing the processed pixels per image from 45.44\,MP to 4.46\,MP---over a 10$\times$ reduction.

\section{DISCUSSION}

For blade 1, COLMAP was unable to produce a sensible sparse reconstruction, despite trying various inlier and filtering settings. While Metashape produced a reasonable reconstruction, our method outperformed it across all three metrics. For blade 2, our reconstruction likewise outperformed both COLMAP and Metashape across all evaluated metrics. The low reprojection error combined with a large number of points is a result of the robust dense matcher, while the longer mean track length is due to our strategy of sampling overlapping image sets to extract keypoint tracks.

Fig.~\ref{fig:dense_model_comparison} highlights the improved completeness of our Gaussian Splatting-based dense model compared to COLMAP and Metashape. Unlike the others, our reconstruction has no gaps; COLMAP is noisy, while Metashape is oversmoothed\footnote{Despite the smoothing setting being disabled.} in textured areas such as erosion or coating peeling. By contrast, our model preserves clear depth variations even with relatively minor blade damage.

Fig. \ref{fig:dense_ablation} shows the following:
(1) \textbf{Full-resolution processing} via tiling greatly improves detail and accuracy.  
(2) \textbf{Background masking} is essential, especially with tiling, as residual background regions produce floating Gaussians. Moreover, soft masks reduce boundary artifacts versus hard masks (which may be caused by inaccurate masks but also by minor inconsistencies between masks of different images or small pose errors).  
(3) \textbf{Adaptive densification} enhances the leading edge, focusing detail where needed, preserving fine textures and yielding less noisy yet detailed depth maps. While the visual renders appear similar, the geometry  reveals smoother, more plausible surfaces, while still preserving depth variations at defect locations.

While representing a significant advance for practical 3D vision in wind blade inspections, our method has limitations. Artifacts remain near the trailing edge, mainly due to the flight path design and limited view diversity. Although the reconstructions show high detail at the leading edge (Figs.~\ref{fig:dense_model_comparison} and \ref{fig:surface_detail}), some features appear to be missing (e.g., some boundaries of coating peeling in Fig. \ref{fig:surface_detail}). Moreover, in the absence of 3D ground-truth data, the reconstruction accuracy cannot be precisely quantified.


\section{CONCLUSION}

We developed a custom data acquisition and scalable 3D reconstruction pipeline tailored to high-resolution UAV-based wind blade inspections. By adopting a coarse-to-fine strategy for efficient overlap detection, our method enables dense keypoint matching at full resolution within practical computational limits and outperforms leading 3D reconstruction software across several metrics. For dense reconstruction, we extend 2D Gaussian Splatting with soft masking, tiled rendering, and blade-specific densification. The resulting models surpass classical MVS in completeness and produce more plausible and detailed surface geometry.

Future work could explore optimized flight paths that balance efficiency and viewpoint diversity for improved 3D reconstruction. Also, motion- and lens-aware deblurring modules tailored to the UAV's dynamics and optics could be developed, as off-the-shelf models had little effect on our nearly sharp images. Combined with multi-view super-resolution before dense reconstruction, this could further enhance surface detail.


\section*{ACKNOWLEDGEMENTS}
Funding: This work was supported by the Energy Transition Fund (ETF) of the Belgian Federal Government (BLEEPID-project; Federal Public Service Economy Belgium).


\bibliographystyle{apalike}
{\small
	\bibliography{references}
}

\end{document}